\documentclass[runningheads]{llncs}
\usepackage{graphicx}
\usepackage{tikz}
\usepackage{comment}
\usepackage{amsmath,amssymb} % define this before the line numbering.
\usepackage{color}

\usepackage{hyperref}
\usepackage{url}
\usepackage{kotex}

\usepackage{booktabs}       % professional-quality tables
\usepackage{amsfonts}       % blackboard math symbols
\usepackage{nicefrac}       % compact symbols for 1/2, etc.
\usepackage{microtype}      % microtypography
\usepackage{wrapfig}

\usepackage{graphicx}
\usepackage{xcolor,colortbl}
\usepackage[caption=false,font=footnotesize]{subfig}
\usepackage{epsfig}
\usepackage{bbm}

\usepackage{enumitem}
\usepackage{hyperref}
\usepackage{multirow}
\usepackage{array}
\newcommand{\PreserveBackslash}[1]{\let\temp=\\#1\let\\=\temp}
\newcolumntype{C}[1]{>{\PreserveBackslash\centering}p{#1}}
\newcolumntype{R}[1]{>{\PreserveBackslash\raggedleft}p{#1}}
\newcolumntype{L}[1]{>{\PreserveBackslash\raggedright}p{#1}}

\newcommand{\xbu}{{\bf x}}
\DeclareMathOperator*{\argmax}{arg\,max}
\DeclareMathOperator*{\argmin}{arg\,min}
\newcommand{\R}{\mathbb{R}}
\newcommand{\E}{\mathbb{E}}

\numberwithin{const2}{const}

\usepackage[accsupp]{axessibility}  % Improves PDF readability for those with disabilities.

\begin{document}
% \linenumbers
\pagestyle{headings}
\mainmatter
\def\ECCVSubNumber{7883}  % Insert your submission number here

\title{Towards Accurate Open-Set Recognition via Background-Class Regularization} % Replace with your title

\begin{comment}
\titlerunning{ECCV-22 submission ID \ECCVSubNumber} 
\authorrunning{ECCV-22 submission ID \ECCVSubNumber} 
\author{Anonymous ECCV submission}
\institute{Paper ID \ECCVSubNumber}
\end{comment}
%******************

%\begin{comment}
\titlerunning{Background-Class Regularization for Open-Set Recognition}
\author{Wonwoo Cho\inst{1,2}\and
Jaegul Choo\inst{1,2}}
\authorrunning{W. Cho and J. Choo}
\institute{KAIST AI \and
Letsur Inc.\\
\email{\{wcho,jchoo\}@kaist.ac.kr}}
%\end{comment}
%******************
\maketitle

\begin{abstract}
In open-set recognition (OSR), classifiers should be able to reject unknown-class samples while maintaining high closed-set classification accuracy. To effectively solve the OSR problem, previous studies attempted to limit latent feature space and reject data located outside the limited space via offline analyses, \emph{e.g.}, distance-based feature analyses, or complicated network architectures. To conduct OSR via a simple inference process (without offline analyses) in standard classifier architectures, we use distance-based classifiers instead of conventional Softmax classifiers. Afterwards, we design a background-class regularization strategy, which uses background-class data as surrogates of unknown-class ones during training phase. Specifically, we formulate a novel regularization loss suitable for distance-based classifiers, which reserves sufficiently large class-wise latent feature spaces for known classes and forces background-class samples to be located far away from the limited spaces. Through our extensive experiments, we show that the proposed method provides robust OSR results, while maintaining high closed-set classification accuracy.

%which can contrast known-class and background-class samples.
%in terms of the distance measure used for classification while keeping robust closed-set classification performance.

%However, previous regularization methods have limited OSR performance, since they categorized known-class data into a single group and then aimed to distinguish them from anomalies.
\keywords{Generalized Open-Set Recognition, Distance-Based Classifiers, Background-Class Regularization, Probability of Inclusion}
\end{abstract}

\section{Introduction}

In machine learning (ML), classification algorithms have achieved great success.
Through recent advances in convolutional neural networks, their classification performance already surpassed the human-level performance in image classification~\cite{He:delving}.
However, such algorithms have usually been developed under a \emph{closed-set} assumption, \emph{i.e.}, the class of each test sample is assumed to always belong to one of the pre-defined set of classes.
Although this conventional assumption can be easily violated in real-world applications (classifiers can face unknown-class data), traditional classification algorithms are highly likely to force unknown-class samples to be classified into one of the known classes.
To tackle this problem, the \emph{open-set recognition (OSR)} problem~\cite{Scheirer:openset} aims to properly classify unknown-class samples as ``unknown'' and known-class samples as one of the known classes.

According to the definition of OSR~\cite{Scheirer:openset}, it is required to properly limit the latent feature space of known-class data.
To satisfy the requirement, various OSR methods were developed based on traditional ML models.
Previously, Scheirer~\emph{et al.}~\cite{Scheirer:svm} calibrated the decision scores of support vector machines (SVMs).
Based on the intuition that a large set of data samples of unknown classes can be rejected if those of known classes are accurately modeled, Jain~\emph{et al.}~\cite{Jain:pisvm} proposed $P_I$-SVM, which utilized the statistical modeling of known-class samples located near the decision boundary of SVMs.
Afterwards, it was attempted to solve the OSR problem based on the principle of the nearest neighbors~\cite{PRM:specialized}.
Taking distribution information of data into account, Rudd~\emph{et al.}~\cite{Rudd:extreme} proposed the extreme value machine which utilizes the concept of margin distributions.

Since deep neural networks (DNNs) have robust classification performance by learning high-level representations of data, OSR methods for DNNs have received great attention.
Based on the theoretical foundations studied in traditional ML-based OSR methods, Bendale and Boult~\cite{Bendale:openmax} proposed the first OSR strategy for DNNs called Openmax, which calibrates the output logits of pre-trained Softmax classifiers.
To improve Openmax, Yoshihashi~\emph{et al.}~\cite{Yoshihashi:crosr} proposed the classification-reconstruction learning to make robust latent feature vectors.
Afterwards, Oza and Patel~\cite{Oza:c2ae} proposed to exploit a class-conditioned autoencoder and use its reconstruction error to assess each input sample.
Sun~\emph{et al.}~\cite{sun:cgdl} employed several class-conditioned variational auto-encoders for generative modeling.

Although previous methods applied \emph{offline analyses} to pre-trained Softmax classifiers or employed complicated DNN architectures, they have limited performance since the classifiers were trained solely based on known-class data.
To mitigate the problem, this paper designs an simple and effective open-set classifier in the \emph{generalized OSR setting}, which uses background-class regularization (BCR) at training time.
Despite its effectiveness, BCR has received a little attention in OSR and previous BCR methods~\cite{Dhamija:reducing,Hendrycks:oe,Liu:energy} are insufficient to properly solve the OSR problem.
%To design an effective open-set classifier that can overcome the previous limitations, we propose a novel BCR method suitable for OSR, which uses a distance-based classifier and a novel loss function for regularization.
%In the next sections, we provide more details about preliminaries our proposed method.
In this paper, we denote the infinite label space of all classes as $\mathcal{Y}$ and use the following class categories, whose definition is also provided in~\cite{Dhamija:reducing,Geng:survey}.

\begin{enumerate}[nosep]
  \item[\tiny$\bullet$] \textbf{Known known classes} (KKCs; $\mathcal{K}= \{1,\cdots,C\} \subset \mathcal{Y}$) include distinctly labeled positive classes, where $\mathcal{U}=\mathcal{Y}\setminus \mathcal{K}$ is the entire unknown classes.
  \item[\tiny$\bullet$] \textbf{Known unknown classes} (KUCs; $\mathcal{B}\subset \mathcal{U}$) include background classes, \emph{e.g.}, labeled classes which are not necessarily grouped into a set of KKCs $\mathcal{K}$.
  \item[\tiny$\bullet$] \textbf{Unknown unknown classes} (UUCs; $\mathcal{A}=\mathcal{U}\setminus \mathcal{B}$) represent the rest of $\mathcal{U}$, where UUCs are not available at training time, but occur at inference time.
\end{enumerate}
Also, we denote $\mathcal{D}_{t}$ as a training set consisting of multiple pairs of a KKC data sample and the corresponding class label $y \in \{1,\cdots,C\}$.
$\mathcal{D}_{test}^{k}$ and $\mathcal{D}_{test}^{u}$ are test sets of KKCs and UUCs, respectively. $\mathcal{D}_{b}$ is a background dataset of KUCs.

\section{Preliminary Studies}

\subsection{The Open-Set Recognition Problem}\label{sec2.1}

The OSR problem addresses a classification setting that can face test samples from classes unseen during training (UUCs).
In this setting, open-set classifiers aim to properly classify KKC samples while rejecting UUC ones simultaneously.
A similar problem to OSR is out-of-distribution (OoD) detection~\cite{hendrycks:baseline}, which typically aims to reject data items drawn far away from the training data distribution.
Conventionally, previous studies such as~\cite{hendrycks:baseline,Liang:odin,Lee:calibration,Lee:maha} assumed that OoD samples are drawn from other datasets or can be even noise data.
In this paper, we aim to reject test data whose classes are unknown but related to the training data, which narrows down the scope of conventional OoD detection tasks.

\begin{figure}[t]
%\vspace{-\intextsep}
\centering{%
\subfloat[Latent feature space]{\includegraphics[width=0.27\textwidth, angle=0]{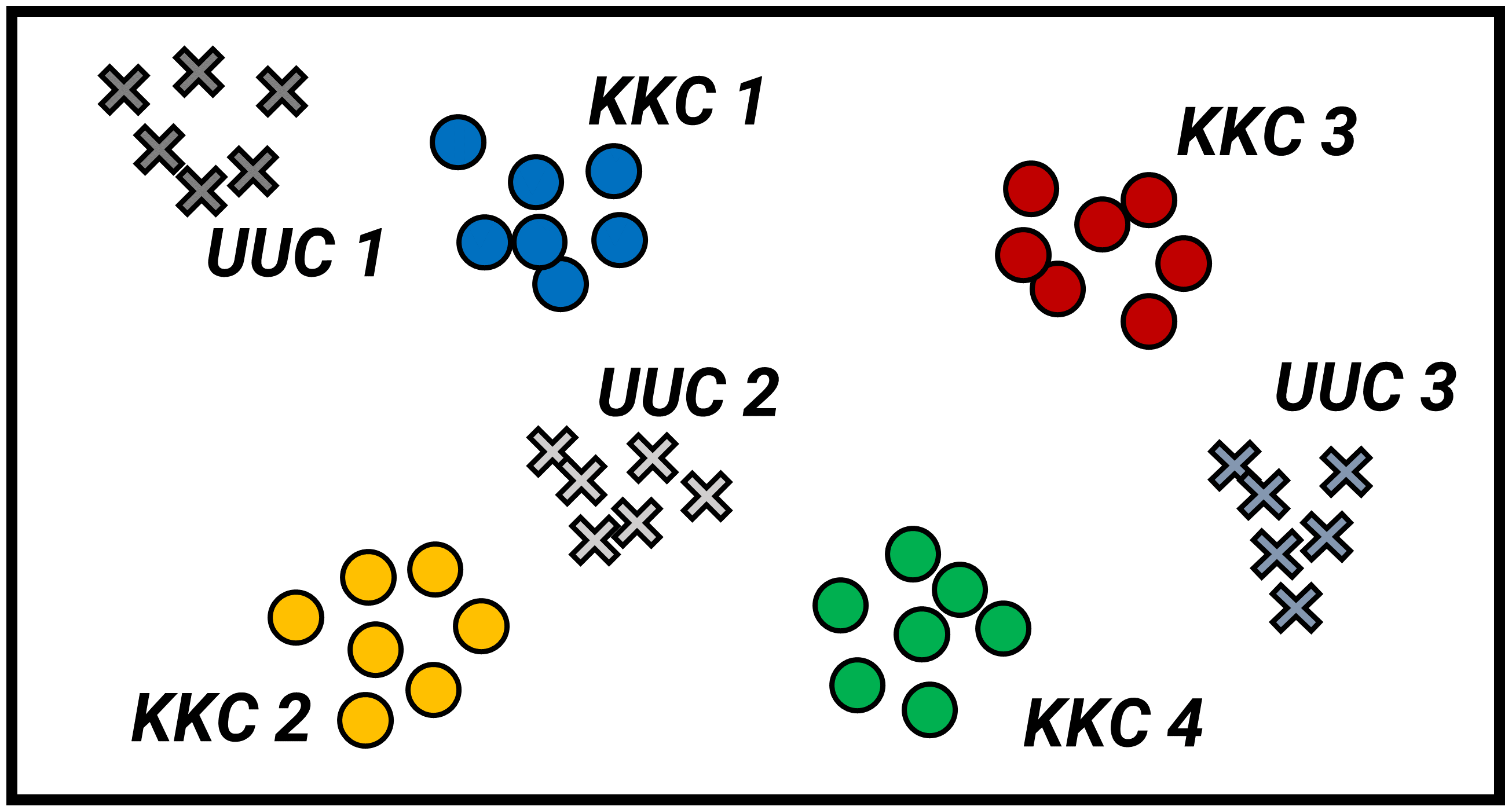}}
\quad
\quad
\subfloat[Closed-set problem]{\includegraphics[width=0.27\textwidth, angle=0]{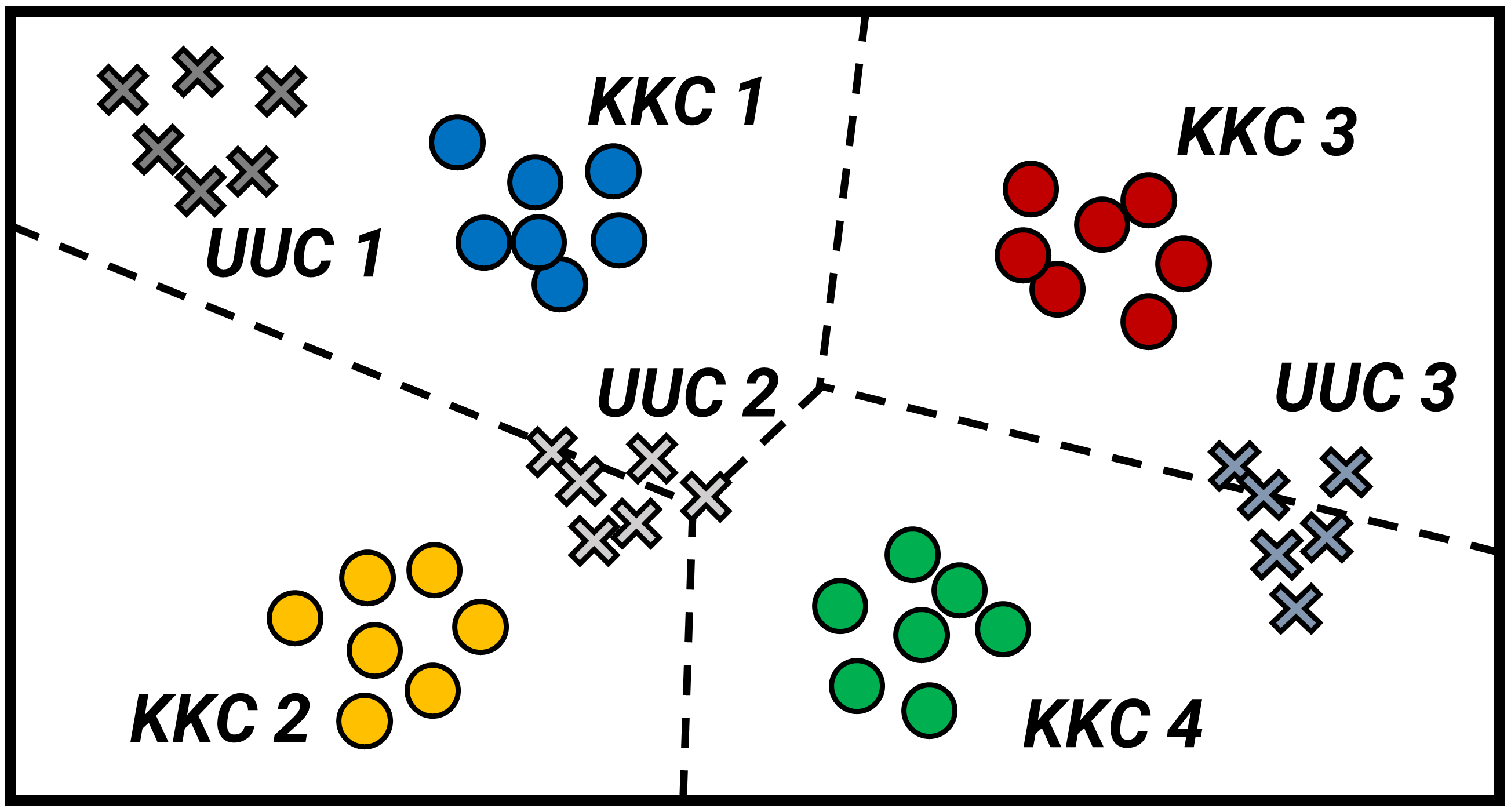}}
\quad
\quad
\subfloat[Open-set problem]{\includegraphics[width=0.27\textwidth, angle=0]{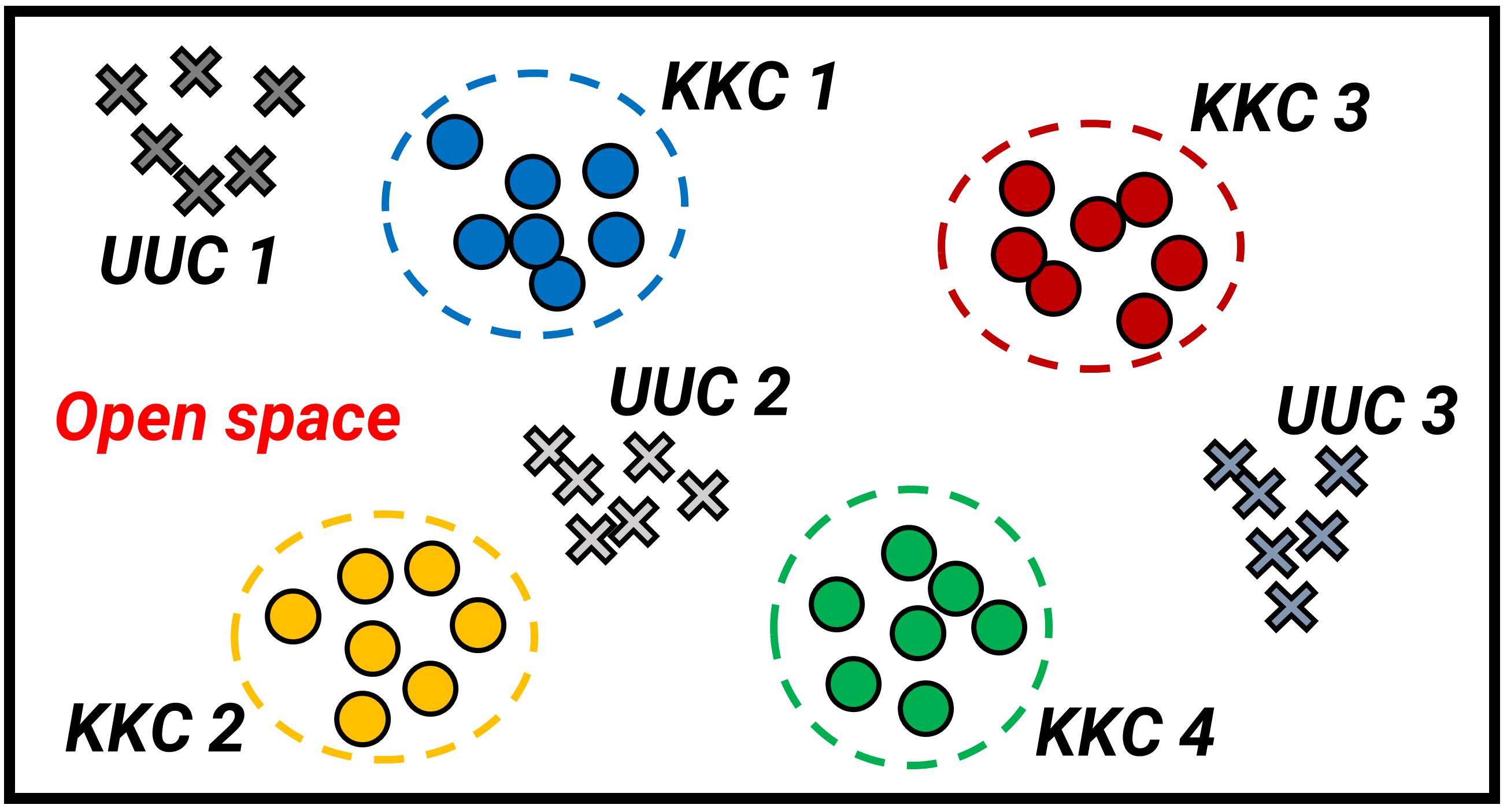}}
}
\caption{Given (a) a latent feature space, we demonstrate (b) closed-set  and (c) open-set problems, where KKCs and UUCs are known and unknown classes, respectively.
}
\label{fig:comparison}
\end{figure}

Previously, Scheirer~\emph{et al.}~\cite{Scheirer:openset} introduced a formal definition of OSR based on the notion of open-space risk $R_{\mathcal{O}}$,
which is a relative measure of a positively labeled union of balls $\mathcal{S}_V$ and
open space $\mathcal{O}$ located far from $\mathcal{S}_V$.
Since labeling any data item in $\mathcal{O}$ incurs open-space risk,
it is straightforward that a classifier cannot be a solution for the OSR problem if the classifier accepts data in infinitely wide regions, \emph{i.e.}, its open-space risk is unbounded ($R_{\mathcal{O}} = \infty$).
The definition implies that essential requirements to solve the OSR problem are 1) \emph{bounding open-space risk} and 2) \emph{ideally balancing it with empirical risk}.

Unlike traditional classifier models, open-set classifiers are required to limit the latent feature space of KKC data to bound their open-space risk.
To ensure open-space risk to be bounded, Scheirer~\emph{et al.}~\cite{Scheirer:svm} introduced compact abating probability (CAP) models.
The principle of CAP models is that if the support region of a classifier decays in all directions from the training data, thresholding the region will bound the classifier's open-space risk~\cite{Boult:survey}.
As depicted in Figure~\ref{fig:comparison}, which compares traditional closed-set and open-set classification problems~\cite{Geng:survey}, building proper \emph{class-wise} CAP models is an effective strategy for OSR.

\subsection{Post-Classification Analysis for Pre-Trained Softmax Classifier}\label{sec2.2}

This paper aims to solve the OSR problem solely based on a standard DNN-based classifier architecture $f$ as a latent feature extractor.
Applying a fully-connected layer to $f$, a conventional Softmax classifier computes the posterior probability of an input $\xbu$ belonging to the $c$-th known class by
\begin{equation}\label{eq:ori_prob}
    P_s(y=c|\xbu) = \frac{\exp({\bf w}_c^T f (\xbu)+b_c)}{\sum_{i=1}^C \exp({\bf w}_i^T f (\xbu)+b_i)},
\end{equation}
where $c \in \{1,\cdots,C\}$, $f(\xbu) \in \R^n$ is the latent feature vector of $\xbu$, and ${\bf w}_c$ and $b_c$ are the weight and bias for the $c$-th class, respectively.
For pre-trained Softmax classifiers, Hendrycks and Gimpel~\cite{hendrycks:baseline} proposed a baseline technique to detect anomalous samples, which imposes a threshold on the predictive confidence of Eq.~(\ref{eq:ori_prob}).
When using the baseline approach to solve the OSR problem, one can estimate the class of each KKC sample and recognize UUC data by
\begin{equation}~\label{eq:softmax_inf}
    \widehat{y} =
    \begin{cases}
    \argmax_{c \in \{1,\cdots,C\}} P_s(y=c|\xbu), & \text{if } \max_{c \in \{1,\cdots,C\}}  P_s(y=c|\xbu) \ge \tau, \\
    C+1 \textrm{ (unknown class)}, & \text{otherwise}.
    \end{cases}
\end{equation}
However, Eq.~(\ref{eq:softmax_inf}) cannot formally bound open-space risk
and formulate class-wise CAP models
since it only rejects test data near the decision boundary of classifiers, thus having infinitely wide regions of acceptance~\cite{Boult:survey}.
Therefore, \emph{post-classification analysis} methods using an auxiliary measure other than the Softmax probability are necessary to build auxiliary CAP models in the latent feature space of $f$, where \emph{distance measures} have been widely employed in previous studies~\cite{Bendale:openmax,Lee:maha}.

To build class-wise CAP models, Openmax~\cite{Bendale:openmax} defined radial-basis decaying functions $\{s(\xbu,i)\}_{i=1}^C$, each of which measures the class-belongingness of $\xbu$ for the $c$-th class, in the latent feature space of $f$.
For each $s(\xbu,c)$, the authors employed distance measures between $f(\xbu)$ and an empirical class mean vector ${\boldsymbol \mu}_c$, \emph{e.g.}, $s(\xbu,c) = D_E^2(f(\xbu), {\boldsymbol \mu}_c) = (f(\xbu)-{\boldsymbol \mu}_c)^T(f(\xbu)-{\boldsymbol \mu}_c)$.
To formulate more effective CAP models, they statistically analyzed the distribution of $s(\xbu,c)$ based on the extreme value theory (EVT)~\cite{Scheirer:evt}, which provides a theoretical foundation that the Weibull distribution is suitable for modeling KKC samples located far from the class mean vectors (extreme samples).
To be specific, Openmax fits a Weibull distribution on extreme samples of the $c$-th class having the highest $D_E(f(\xbu), {\boldsymbol \mu}_c)$ values,
where its cumulative distribution function (CDF) formulates the \emph{probability of inclusion} $P_I(\xbu,c)$~\cite{Jain:pisvm,Rudd:extreme}, \emph{i.e.}, $P_I(\xbu,c) = 1 - \texttt{WeibullCDF}$, which rapidly decays near the extreme samples.
Based on $P_I(\xbu,c)$, the decision rule of
Eq.~(\ref{eq:softmax_inf}) can be calibrated to conduct OSR with Softmax classifiers.

%$P_I(\xbu,k)$ can make an effective bound $\mathcal{R}_k$ by selecting a threshold between $0$ and $1$.

\subsection{Background-Class Regularization}\label{sec2.3}

Although they need additional inference procedures (\emph{e.g.}, EVT modeling), previous offline analyses may have limited OSR performance since the classifiers were trained solely based on known-class data.
%Also, the Softmax classifiers are likely to yield inaccurate latent feature space for the post-classification analyses using distance measures.
To obtain robust empirical results without complicated analyses, one can use the strategy of BCR at the training phase, which exploits background-class (KUC) samples as surrogates of UUC data.
Geng~\emph{et al.}~\cite{Geng:survey} argued that the generalized OSR setting that utilizes KUC samples is still less-explored and an important research direction for robust OSR.
%In the generalized OSR setting, classifiers should consider the following data classes among the infinite label space of all classes $\mathcal{Y}$~\cite{Dhamija:reducing,Geng:survey}.

Conventionally, a loss function for training classifiers with BCR can be 
\begin{equation}\label{eq:previous}
    \mathcal{L} = \mathcal{L}_{cf} + \lambda \mathcal{L}_{bg} = \E_{(\xbu^k, y)\sim\mathcal{D}_{t}} \left[- \log P_s(y|\xbu^k) + \lambda \E_{\xbu^b \sim\mathcal{D}_{b}}\left[f_{reg}\left(\xbu^k, y, \xbu^b \right)  \right]\right],
\end{equation}
where $\mathcal{L}_{cf}$ and $\mathcal{L}_{bg}$ are the loss terms for closed-set classification and BCR, respectively, and $\lambda$ is a hyperparameter.
For $\mathcal{L}_{bg}$, previous studies designed their own $f_{reg}$, where \cite{Dhamija:reducing} proposed the objectosphere loss for OSR, and \cite{Hendrycks:oe} and \cite{Liu:energy} employed the uniformity and the energy losses for OoD detection, respectively.

In this paper, we tackle the following limitations of the previous BCR methods.
\begin{enumerate}[nosep]
  \item[\tiny$\bullet$] In the previous BCR methods, $\mathcal{L}_{bg}$ were designed to make normal data and anomalies more distinguishable in terms of the corresponding anomaly scores.
  Since they categorized normal data into a single group (did not consider the classes) in $\mathcal{L}_{bg}$,
  the previous methods may have limited performance in rejecting UUC data and maintaining robust closed-set classification results.
  \item[\tiny$\bullet$] The previous methods using the decision rule of Eq.~(\ref{eq:softmax_inf}) (\emph{e.g.}, objectosphere~\cite{Dhamija:reducing} and uniformity~\cite{Hendrycks:oe}) cannot bound open-space risk.
  Although one can use post-classification analyses to bound open-space risk, trained latent feature space can be
  inappropriate for using another metric such as distance measures.
  \item[\tiny$\bullet$] To increase the gap between KKC and KUC data
  in terms of latent feature magnitude and energy in the objectosphere~\cite{Dhamija:reducing} and the energy~\cite{Liu:energy} losses, respectively,
  it is necessary to find proper margin parameters for each dataset.
\end{enumerate}

\section{Proposed Method}

\subsection{Overview}\label{sec2.4}

Using a standard classifier $f$, this paper aims to design open-set classifiers having simple yet effective inference steps.
% by using the principle of BCR.
%To overcome the limitations described in Section~\ref{sec2.3}, we use distance-based classifiers and propose a novel BCR strategy based on the framework of Eq.~(\ref{eq:previous}).
In the following, we summarize our method.
\begin{enumerate}[nosep]
  \item[\tiny$\bullet$] 
  Instead of applying fully-connected layers to feature extractors $f$, we use the principle of linear discriminant analysis (LDA)~\cite{Murphy:machine} to classify images based on a distance measure.
  By simply imposing a threshold on the distance as in Eq.~(\ref{eq:softmax_inf}),
  our classifiers can easily build class-wise CAP models. (Section~\ref{sec3.1})
  %without additional offline analyses.
  \item[\tiny$\bullet$] Afterwards, we propose a novel BCR strategy suitable for the distance-based classifiers. Following the convention of Eq.~(\ref{eq:previous}), we design our own $\mathcal{L}_{bg}$ called \emph{class-inclusion loss}, where our total loss is function defined by
  \begin{equation}~\label{eq:totloss}
    \mathcal{L} = \mathcal{L}_{cf} + \lambda \mathcal{L}_{bg} = \mathcal{L}_{cf} + \lambda ( \mathcal{L}_{bg,k}+\mathcal{L}_{bg,u})
  \end{equation}
  %For $\mathcal{L}_{bg}$, we design a loss function that does not require data-dependent margin parameters.
  The class-inclusion loss first limits the feature space of KKC data by formulating \emph{explicit} class-wise boundaries, and then forces KUC data to be located outside the boundaries at each training iteration.
  Our loss is designed to increase the distance gaps between KKC and KUC samples while maintaining robust closed-set classification performance. (Sections~\ref{sec3.2} and~\ref{sec3.3})
\end{enumerate}
For a better understanding of the training and inference processes of our method, we provide their detailed algorithm in our supplementary materials.

\subsection{Distance-Based Classification Models}\label{sec3.1}

\subsubsection{Distance-based classifiers.}

To train a robust open-set classifier, we formulate a \emph{distance-based classifier}
as an alternative of Eq.~(\ref{eq:ori_prob}):
\begin{equation}~\label{eq:disc}
    P_d(y=c|\xbu) = \frac{P_c\cdot \mathcal{N}(f(\xbu)|{\boldsymbol \mu}_c, {\bf I})}{\sum_{i=1}^C P_i \cdot \mathcal{N}(f(\xbu)|{\boldsymbol \mu}_i, {\bf I})} = \frac{P_c \cdot \exp\left(-D_E^2(f(\xbu), {\boldsymbol \mu}_c)\right)}{\sum_{i=1}^C P_i \cdot \exp\left(-D_E^2(f(\xbu), {\boldsymbol \mu}_i)\right)}
\end{equation}
where Eq.~(\ref{eq:disc}) uses the principle of LDA and $\mathcal{L}_{cf} = \E_{(\xbu^k, y)\sim\mathcal{D}_{t}} [- \log P_d(y|\xbu^k)]$.
In Eq.~(\ref{eq:disc}), we exploit an identity covariance matrix ${\bf I}$ and $P_c = P(y=c)=C^{-1}$ for all $c$ for KKCs.
The classifier estimates the class of each $\xbu$ via $D_E^2(f(\xbu), {\boldsymbol \mu}_c) = (f(\xbu)-{\boldsymbol \mu}_c)^T(f(\xbu)-{\boldsymbol \mu}_c)$, the Euclidean distance between $f(\xbu) \in \R^n$ and ${\boldsymbol \mu}_c  \in \R^n$, where we call ${\boldsymbol \mu}_c$ a \emph{class-wise anchor}.
To ensure sufficiently large distance gaps between the pairs of initial class-wise anchors,
we randomly sample each $\boldsymbol{\mu}_c$ from the standard Gaussian distribution and then set each $\boldsymbol{\mu}_c$ as a trainable vector.
For distance analysis results of such randomly sampled vectors,
see~\cite{Izmailov:semi}.

%Instead of updating or computing empirical ${\boldsymbol \mu}_c$,
%we  then fix it as an anchor during the training process.

%Such strategy also showed successful results in~\cite{Izmailov:semi}, which formulated \emph{generative classifiers} based on the principle of Gaussian mixture models.

%\vspace{-1\baselineskip}
\subsubsection{Decision rule.}

At inference time, each KKC sample $\xbu$ can be classified via $\widehat{y} = \argmin_{c \in \{1,\cdots,C\}} D_E^2(f(\xbu), {\boldsymbol \mu}_c)$.
Furthermore, applying a threshold to $D_E^2(f(\xbu), {\boldsymbol \mu}_c)$ can bound open-space risk by formulating class-wise CAP models
%Therefore, one can properly conduct OSR without additional post-classification analyses
as follows:
\begin{equation}~\label{eq:inference}
    \widehat{y} =
    \begin{cases}
    \argmin_{c \in \{1,\cdots,C\}} D_E^2(f(\xbu), {\boldsymbol \mu}_c), & \text{if } \max_{c \in \{1,\cdots,C\}}  - D_E^2(f(\xbu), {\boldsymbol \mu}_c) \ge \tau, \\
    C+1 \textrm{ (unknown class)}, & \text{otherwise}.
    \end{cases}
\end{equation}
As Eq.~(\ref{eq:inference}) employs the same metric $D_E$ for classification and UUC rejection, our method may support more accurate latent feature space analysis for OSR than the previous OSR methods using post-classification analyses.

The concept of distance-based classification was also employed in prototypical networks~\cite{snell:prototypical}, nearest class mean classifiers~\cite{Mensink:ncm}, and the previous studies of the
center loss function~\cite{wen:centerloss} and convolutional prototype classifiers~\cite{yang:convolutionalp}.
In addition, polyhedral conic classifiers~\cite{cevikalp:polyhedral}
used the idea of returning compact class regions for KKC samples based on distance-based feature analyses.
It is noteworthy that our main contribution is a novel BCR method that can effectively utilize KUC samples in a distance-based classification scheme (described in Sections~\ref{sec3.2} and~\ref{sec3.3}),
not the distance-based classifier method itself.
To the best of our knowledge, we are the first to discuss the necessity of distance-based BCR methods for OSR and propose a reasonable regularization method for distance-based classifiers.

%By sampling pairs of a KKC data item and the corresponding label $(\xbu, y)$ in $\mathcal{D}_{train}$,
%we can reduce empirical risk by using a loss function $\mathcal{L}_{cf}=\texttt{NLL}({\bf p}(\xbu), y)$, which is
%the negative log-likelihood loss~\cite{Goodfellow:dlbook},
%where ${\bf p}(\xbu)[k] = P(y=k|\xbu)$ in Eq.~\eqref{eq:disc} for $k \in \{1,\cdots,K\}$.

\subsection{Background Class Regularization for Distance-based Classifiers}\label{sec3.2}

\subsubsection{Intuition and hypersphere classifiers.}

To obtain robust OSR performance via Eq.~(\ref{eq:inference}), we aim to design a BCR method suitable for distance-based classifiers, which uses $\mathcal{D}_{t}$ and $\mathcal{D}_b$ as surrogates of $\mathcal{D}_{test}^k$ and $\mathcal{D}_{test}^u$ at training time, respectively.
Although it cannot provide any information of $\mathcal{D}_{test}^{u}$, $\mathcal{D}_b$ can be effective to limit the latent feature space of KKCs, while reserving space for UUCs.
With $\mathcal{D}_b$, it is intuitive that the primary objective of BCR for Eq.~(\ref{eq:inference}) is to make KUC samples located far away from ${\boldsymbol \mu}_i$ for all classes $i \in \{1,\cdots,C\}$.

Before we illustrate our BCR method, we first introduce hypersphere classifiers (HSCs)~\cite{ruff:hsc}.
%, whose concept was also used in \cite{liznerski:fcdd}.
An HSC conducts anomaly detection by using a feature extractor $g$, where its anomaly score for an input $\xbu$ is the Euclidean distance between a single center vector $\boldsymbol{\mu}$ and $g(\xbu)$.
When training the HSC model, the authors used normal and background data, $\mathcal{D}_{t}$ and $\mathcal{D}_{b}$, respectively, and a loss function
\begin{equation}\label{eq:hscobj}
    \E_{\xbu^k\sim\mathcal{D}_{t}} \left[h\left(D_E^2\left(g(\xbu^k), {\boldsymbol \mu}\right)\right)\right] - \E_{\xbu^b \sim\mathcal{D}_{b}}\left[ \log\left(1-\exp\left(-h\left(D_E^2\left(g(\xbu^b), {\boldsymbol \mu}\right)\right)\right)\right) \right].
\end{equation}
The loss function is designed to decrease the Euclidean distances between normal samples $\xbu^k$ and $\boldsymbol{\mu}$ while increasing the distances for background samples $\xbu^b$.
In Eq.~(\ref{eq:hscobj}), $h(x) = \sqrt{x + 1} - 1$, which implies that the Euclidean distance  $D_E^2(g(\xbu), {\boldsymbol \mu})$ is scaled into the range of $(0,1]$ via $\exp(-h(D_E^2(g(\xbu), {\boldsymbol \mu})))$.

%\vspace{-1\baselineskip}
\subsubsection{Background-class regularization strategy.}

It is straightforward that the decision rule of Eq.~(\ref{eq:inference}) employs the principle of HSCs in a class-wise manner.
%, where the original HSC formulates its decision boundary for anomaly detection as a hypersphere having a constant radius.
In other words, the class-wise HSC for the $c$-th class determines whether a test sample belongs to the $c$-th class by computing $D_E^2(f(\xbu), {\boldsymbol \mu}_c)$, where the input is determined as UUC  if the entire class-wise HSCs reject the data item.
Thus,
%unlike the original training objective of HSC,
a proper BCR strategy for distance-based classifiers
%should make KKC data to be accepted by the corresponding class-wise HSC (decrease $D_E^2(f(\xbu^k), {\boldsymbol \mu}_c)$),
should force each KUC sample $\xbu^b$ to be rejected by the entire class-wise HSCs (increase $D_E^2(f(\xbu^b), {\boldsymbol \mu}_i)$ for all $i$).
Since it is inefficient to consider the entire KKCs to regularize $f$ with $\xbu^b$ at each iteration, we approximate the process by only taking 
the \emph{closest} class-wise HSC into account (increase $\min_{i\in\{1,\cdots,C\}} D_E^2(f(\xbu^b), {\boldsymbol \mu}_i)$).

Although one can adopt Eq.~(\ref{eq:hscobj}) to formulate $\mathcal{L}_{bg}$ for distance-based classifiers, scaling $D_E^2(f(\xbu), {\boldsymbol \mu}_c)$ into $(0,1]$ via $\exp(-h(D_E^2(f(\xbu), {\boldsymbol \mu}_c)))$, which rapidly decays near ${\boldsymbol \mu}_c$, can be insufficient to move KUC data far away from class-wise anchors.
Therefore, we design $\mathcal{L}_{bg}$ that can
guarantee sufficient spaces for KKC data and simultaneously force KUC samples located outside the limited class-wise spaces.

\subsection{Probability of Inclusion and Class-Inclusion Loss}\label{sec3.3}

As we described in Section~\ref{sec2.2},
the probability of inclusion builds effective CAP models, since it is
designed to rapidly decay near extreme data, \emph{i.e.}, $P_I(\xbu,c) \approx 1$
in the region that a majority of class-$c$ KKC samples are located.
In the following, we introduce a novel regularization method for distance-based classifiers based on the principle of the probability of inclusion, and then design a loss function.

%\vspace{-1\baselineskip}
\subsubsection{Probability of inclusion for distance-based classifiers.}

For pre-trained Softmax classifiers, Openmax~\cite{Bendale:openmax} formulated the probability of inclusion via EVT modeling at inference time, where the strategy is to find \emph{implicit} class-wise boundaries that distinguish KKCs from UUCs.
%If such implicit boundaries are available at training time, it is intuitive to force KUC samples to be located outside the boundaries.
However, such EVT-based analysis can be intractable at each training iteration,
since it requires computationally-expensive and parameter-sensitive processes.
In addition, it is inappropriate to make boundaries by analyzing features which are not properly trained yet.

Thus, we build \emph{explicit} class-wise boundaries by formulating $P_I(\xbu,c)$ based on the underlying assumption of LDA, and then use the boundaries for regularization without additional analysis of latent feature distribution.
%since we define the training and the inference processes of distance-based classifiers based on the principle of LDA.
Under the assumption of LDA that each class-$c$ latent feature vector is drawn from a unimodal Gaussian distribution $\mathcal{N}(f(\xbu)|{\boldsymbol \mu}_c, {\bf I})$, the Euclidean distance $D_E^2(f(\xbu), {\boldsymbol \mu}_c)$, a simplified version of the Mahalanobis distance, can be assumed to follow the Chi-square distribution having the degree of freedom $n$.
Then, we have
\begin{equation}~\label{eq:chi2pdf}
    P\left(D_E^2(f(\xbu), {\boldsymbol \mu}_c)=t\right) = \frac{t^{\frac{n}{2}-1}}{2^{\frac{n}{2}}\cdot\Gamma(n/2)}\cdot \exp\left(-\frac{t}{2}\right),
\end{equation}
where $t \ge 0$, $\Gamma(\cdot)$ is the Gamma function, and $n$ is the dimension of $f(\xbu)$.

As previous studies~\cite{Jain:pisvm,Bendale:openmax,Rudd:extreme} formulated the probability of inclusion by computing the CDF of the Weibull distribution, \emph{i.e.}, $P_I(\xbu,c) = 1 - \texttt{WeibullCDF}$, we define our $P_I(\xbu,c)$ by using the CDF of Eq.~(\ref{eq:chi2pdf}) as follows:
\begin{equation}~\label{eq:chi2cdf}
    P_I(\xbu,c)  = 1- \int_0^{{D_E^2(\xbu, c)}/{2}}\frac{t^{n/2-1}}{\Gamma(n/2)}\cdot\exp\left(-t\right) dt = \frac{\Gamma(n/2,D_E^2(f(\xbu), {\boldsymbol \mu}_c)/2)}{\Gamma(n/2)},
\end{equation}
where $\Gamma(\cdot, \cdot)$ is the upper incomplete Gamma function. It is noteworthy that Eq.~(\ref{eq:chi2cdf}) can be easily computable via $\texttt{igammac}$ function in PyTorch~\cite{pytorch}.

%\vspace{-1\baselineskip}
\subsubsection{Class-inclusion loss function.}

\begin{wrapfigure}[9]{r}{0.35\textwidth}
    \centering
    \vspace{-\intextsep}
    \hspace*{-.75\columnsep}
    {\includegraphics[width=0.35\textwidth]{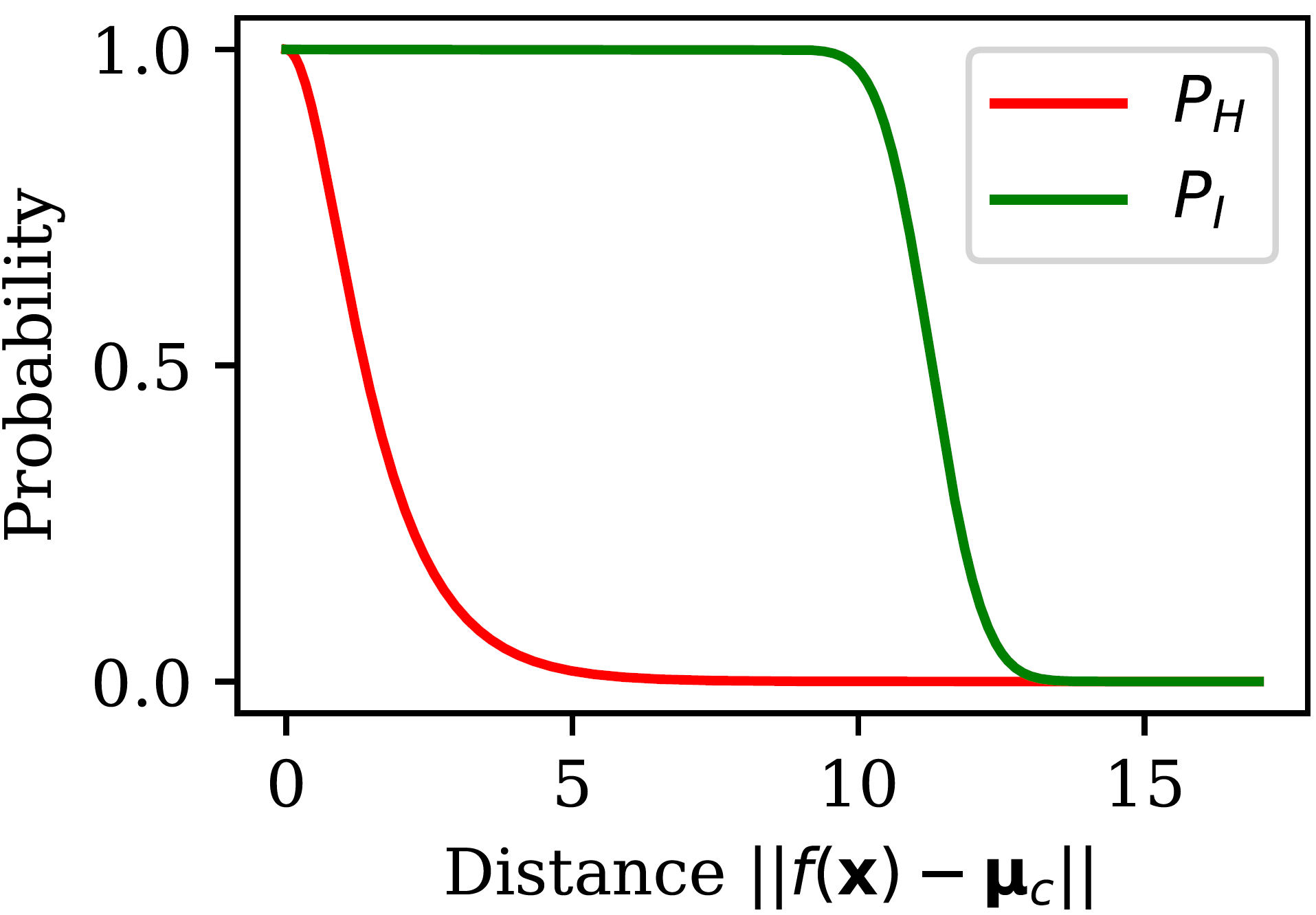}\vspace{-0.3\intextsep}}
    \caption{$P_H$ and $P_I$ (Ours)}
    \flushbottom
    \label{fig:probinc}
\end{wrapfigure}

Based on $\mathcal{D}_{t}$, $\mathcal{D}_{b}$, and our $P_I(\xbu,c)$ of Eq.~(\ref{eq:chi2cdf}),
the primary objective of the proposed BCR strategy, which aims to force each KUC data sample to be located far away from the closest class-wise HSC, can be achieved by employing a loss function $\mathcal{L}_{bg,u} = \E_{\xbu^b \sim\mathcal{D}_{b}}[-\log(1 - \max_{i\in\{1,\cdots,C\}} P_I(\xbu^b,i)) ]$.
To compare $P_I(\xbu,c)$ and $P_H(\xbu,c) = \exp(-h(D_E^2(f(\xbu), {\boldsymbol \mu}_c)))$, which was used in Eq.~(\ref{eq:hscobj}),
we plot $P_I(\xbu,c)$ and $P_H(\xbu,c)$ in Figure~\ref{fig:probinc} with respect to $||f(\xbu) - {\boldsymbol \mu}_c||$ by assuming $n=128$.
The figure implies that unlike $P_H(\xbu,c)$, our $P_I(\xbu,c)$ can
assign sufficiently large space for KKC data and force KUC samples to be located outside the space.
Also, it is noteworthy that our regularization method based on $P_I(\xbu,c)$
does not require any margin parameters dependent on datasets or the dimension of latent features.

At training time, $P_I(\xbu,c) = 0.5$ constructs an auxiliary decision boundary between the $c$-th class KKC data and the other data items, where $\mathcal{L}_{bg,u}$ makes a majority of KUC data to be located outside the entire class-wise boundaries.
However, $\mathcal{L}_{bg,u}$ can be insufficient to achieve robust UUC rejection and closed-set classification results, since it does not control correctly classified KKC samples to be located inside the corresponding class-wise boundaries.
Therefore, in addition to $\mathcal{L}_{cf} = \E_{(\xbu^k, y)\sim\mathcal{D}_{t}} [- \log P_d(y|\xbu^k)]$, we apply another loss $\mathcal{L}_{bg,k}$ to KKC data to \emph{maintain} high closed-set classification accuracy and enhance the gap between KKC and KUC samples in terms of the Euclidean distance.
By formulating $\mathcal{L}_{bg,k} = \E_{(\xbu^k,y) \sim\mathcal{D}_{t}}[-\mathbbm{1}(y = \hat{c})\log(P_I(\xbu^k,\hat{c}))]$,
where $\hat{c} = \argmax_{i\in\{1,\cdots,C\}} P_I(\xbu^k,i)$,
we define our $\mathcal{L}_{bg}$ as $\mathcal{L}_{bg,k} + \mathcal{L}_{bg,u}$ and call $\mathcal{L}_{bg}$ the \emph{class-inclusion loss}.

In our total loss (Eq.~(\ref{eq:totloss})), $\mathcal{L}_{cf}$ makes KKC samples be correctly classified, $\mathcal{L}_{bg,u}$ makes KUC samples located outside the explicit class-wise boundaries, and $\mathcal{L}_{bg,k}$ additionally regularizes correctly classified KKC samples.
It is noteworthy that we use an additional loss for KKC samples after they are correctly classified, to prevent obstructions in training closed-set classifiers at early iterations.

\section{Experiments}\label{sec4}

Through extensive experiments, we compared our class-inclusion loss for distance-based classifiers to the objectosphere~\cite{Dhamija:reducing},
the uniformity (also widely known as OE)~\cite{Hendrycks:oe}, and the energy~\cite{Liu:energy} losses for conventional Softmax classifiers.
This section aims to show that whether our approach provides competitive UUC rejection results, while keeping high closed-set classification accuracy.
Furthermore, we conducted additional experiments and provided the corresponding discussions.

\subsection{Experimental Settings}\label{sec4.1}

For evaluation, we first measured the closed-set classification accuracy.
To quantify the accuracy of UUC data rejection,
we also measured the area under the receiver operating characteristic curve (AUROC).
Also, we used the open-set classification rate (OSCR) as additional OSR accuracy measure by quantifying the correct closed-set classification rate when the false positive rate for UUC rejection is $10^{-1}$.
For in-depth details of OSCR, see~\cite{Dhamija:reducing}.
As $\mathcal{D}_b$, we used ImageNet~\cite{Olga:imagenet}, which was also employed in~\cite{Li:background}.
To ensure that the classes of $\mathcal{D}_b$ and our test sets are disjoint, we used only the remaining classes of ImageNet, which are not included in the test sets.
In our experiments, we considered the following two settings.

%\vspace{-1\baselineskip}
\subsubsection{Setting 1.}\label{sec4.1.1}

In Setting 1, a single dataset was split into KKCs and UUCs, where we used the KKCs in the training set as $\mathcal{D}_{t}$, and the KKCs and UUCs in the test set as $\mathcal{D}_{test}^k$ and $\mathcal{D}_{test}^u$, respectively.
Following the protocol in~\cite{Neal:counterfactual}, which were also employed in~\cite{Oza:c2ae,sun:cgdl}, we conducted experiments by using the following standard datasets: SVHN~\cite{Netzer:svhn}, CIFAR10 \& CIFAR100~\cite{Krizhevsky:cifar}, and TinyImageNet~\cite{Le:tinyimagenet}.
%In what follows, we describe KKCs and UUCs in each experiment.
\vspace{0.4\baselineskip}\\
\emph{SVHN, CIFAR10} \quad For SVHN and CIFAR10, each of which consists of images of 10 classes, each dataset was randomly partitioned into 6 KKCs and 4 UUCs.
\vspace{0.4\baselineskip}\\
\emph{CIFAR+10, CIFAR+50} \quad For CIFAR+$M$, we employed randomly selected 4 classes of CIFAR10 as KKCs and $M$ classes of CIFAR100 as UUCs.
\vspace{0.4\baselineskip}\\
\emph{TinyImageNet} \quad For a larger number of classes, we randomly selected 20 classes of TinyImageNet as KKCs and then used the remaining 180 classes as UUCs.

%\vspace{-1\baselineskip}
\subsubsection{Setting 2.}\label{sec4.1.2}

By using the training and the test sets of a single dataset as $\mathcal{D}_{t}$ and $\mathcal{D}_{test}^k$, respectively, we employed the test set of another dataset relatively close to $\mathcal{D}_{t}$ as $\mathcal{D}_{test}^u$ in Setting 2.
Adopting the experiment settings in \cite{Yoshihashi:crosr} and \cite{Liang:odin},
we used the entire classes of a dataset as KKCs for CIFAR10 \& CIFAR100.
For UUC dataset, TinyImageNet, LSUN~\cite{Yu:lsun}, and iSUN~\cite{Xu:isun} were selected. 
TinyImageNet and LSUN consists of 10,000 test samples each, where the samples in each dataset were resized (R) or cropped (C) into the size $32 \times 32$.
The iSUN dataset has 8,925 test samples and they were also resized into the size of $32 \times 32$.
The modified datasets can be obtained in the Github repository of~\cite{Liang:odin}.

\subsection{Training Details}\label{sec4.2}

\subsubsection{Network selection.}

For $f$, we employed the Wide-ResNet (WRN)~\cite{Zagoruyko:wrn} and then used its penultimate layer $f(\xbu) \in \R^n$ for the latent feature vector of each input sample $\xbu$.
For CIFAR10 and TinyImageNet, we used WRN 40-2 with a dropout rate of $0.3$,
where WRN 28-10 was employed for CIFAR100 with the same dropout rate.
For SVHN, we used WRN 16-4 with a dropout rate of $0.4$.
Such network selection was determined by referring the experiments in~\cite{Hendrycks:oe,Zagoruyko:wrn}.

%\vspace{-1\baselineskip}
\subsubsection{Parameters.}
For the entire BCR methods,
we set the mini-batch sizes of KKC training samples and KUC samples to $128$.
We kept $\lambda$ as a constant during training, \emph{i.e.}, each $f$ was trained with the BCR method \emph{from scratch}.
To select hyperparameters and margin parameters of the previous regularization methods,
we followed the official implementations\footnote{\url{https://github.com/Vastlab/Reducing-Network-Agnostophobia}}$^{,}$\footnote{\url{https://github.com/hendrycks/outlier-exposure}}$^{,}$\footnote{\url{https://github.com/wetliu/energy_ood}}.
For SVHN, CIFAR10, CIFAR100, and TinyImageNet, we trained the corresponding classifiers for $80$, $100$, $200$, and $200$ epochs, respectively, where we used the stochastic gradient descent for optimization.
For SVHN and the other datasets, we used initial learning rates of 0.01 and 0.1, respectively, and a cosine learning rate decay~\cite{Loshchilov:sgdr}.
We also used the learning rate warm-up strategy for the first 5 epochs of each training process.

\subsection{Results}\label{sec:4.3}

The OSR results of our proposed approach and the previous methods are reported in Tables~\ref{tb:table1} and~\ref{tb:table2}.
All the reported values were averaged over five randomized trials, by randomly sampling seeds, data splits of KKCs and UUCs, and class-wise anchors.
In the tables, $\uparrow$ and $\downarrow$ indicate higher-better and lower-better measures, respectively,
where underlined values present the best scores.

\begin{table}[!t]
\caption{Comparison with the previous BCR methods in the first setting.}
%\vspace{0.5\intextsep}
\label{tb:table1}
\centering
\resizebox{1\textwidth}{!}{
\begin{tabular}{cC{1.75in}C{1.75in}C{1.75in}}
\toprule
\multirow{2.5}{*}{Experiments}  & Accuracy ($\uparrow$) & AUROC ($\uparrow$) & OSCR ($\uparrow$)\\
\cmidrule(lr){2-2}\cmidrule(lr){3-3}\cmidrule(lr){4-4}
 & \multicolumn{3}{c}{Objectosphere / Uniformity / Energy / Class-inclusion (Ours)}\\
\midrule
\midrule
SVHN & 0.968 / 0.966 / 0.972 / \underline{0.974} & 0.935 / 0.927 / 0.911 / \underline{0.956} & 0.813 / 0.793 / 0.774 / \underline{0.854}\\
CIFAR10 & 0.964 / 0.964 / 0.956 / \underline{0.973} & 0.942 / 0.923 /  0.933 / \underline{0.948} & 0.851 / 0.814 / 0.807 / \underline{0.870} \\
CIFAR+10 & \multirow{2}{*}{0.958 / 0.969 / 0.949 / \underline{0.976}} & 0.945 / 0.950 / 0.936 / \underline{0.961} & 0.839 / 0.867 / 0.808 / \underline{0.881} \\
CIFAR+50 &   & 0.944 / 0.942 / 0.937 / \underline{0.957} & 0.837 / 0.837 / 0.808 / \underline{0.865} \\
TinyImageNet & 0.778 / 0.779 / 0.715 / \underline{0.802} & 0.755 / 0.771 / 0.727 / \underline{0.785} & 0.484 / 0.488 / 0.357 / \underline{0.493} \\
%\midrule
%\textbf{Average} & 0.917 / 0.920 / 0.898 / \textbf{0.931} & 0.900 / 0.899 / 0.885 / \textbf{0.913} & 0.764 / 0.758 / 0.708 / \textbf{0.787} \\
\bottomrule
\end{tabular}
}
\end{table}

%\vspace{-1\baselineskip}
\subsubsection{Setting 1.}

For the first setting, Table~\ref{tb:table1} compares our proposed BCR methods for distance-based classifiers with the previous approachs designed for Softmax classifiers.
The results demonstrate that our proposed method obtained robust UUC rejection results, which were superior to the results of the previous approaches.
It is noteworthy that our method achieved higher classification accuracy values, which were critical in acquiring better OSR results in terms of the OSCR measure, than the previous methods.
Such results imply that the proposed framework effectively satisfies the two essential requirements described in Section~\ref{sec2.1}, bounding open-space risk and ideally balancing it with empirical risk.

%\vspace{-1\baselineskip}
\subsubsection{Setting 2.}

%For our proposed BCR method and the previous methods, the corresponding OSR results in the second setting of our experiments are reported in .
In Table~\ref{tb:table2}, we present our experiment results of the second setting.
When using the CIFAR10 and CIFAR100 datasets as KKC data, our approach achieved the highest closed-set classification accuracy, which is consistent with the experiment results of Setting 1.
Furthermore, by averaging the AUROC and the OSCR values over the various UUC datasets, the table shows that our model outperformed the previous methods in the second setting.
%In summary, the experiment results of Settings 1 and 2 present that our proposed method %can successfully train a robust open-set classifier.

%\vspace{-1\baselineskip}
\subsubsection{Average runtime.}

We conducted all the experiments with PyTorch and two GeForce RTX 3090 GPUs.
At each trial in the CIFAR10 experiment of Setting 1, the running time of each training epoch took 28 seconds for our method, where its OSR evaluation required approximately 6.5 seconds.
We observed that the other methods take similar running time at their training and inference phases.

\begin{table}[!t]
\caption{Comparison with the previous methods in the second setting.
The corresponding classification accuracy values are reported in the first column.}
\label{tb:table2}
%\vspace{0.5\intextsep}
\centering
\resizebox{1\textwidth}{!}{
\begin{tabular}{ccC{1.75in}C{1.75in}}
\toprule
 \multirow{2.5}{*}{$\mathcal{D}_{t}/\mathcal{D}_{test}^{k}$} & \multirow{2.5}{*}{$\mathcal{D}_{test}^{u}$} & AUROC ($\uparrow$) & OSCR ($\uparrow$) \\
\cmidrule{3-4}
 & & \multicolumn{2}{c}{Objectosphere / Uniformity / Energy / Class-inclusion (Ours)}\\
\midrule
\midrule
\multirow{6.5}{*}{\shortstack{CIFAR10 \\ \vspace{2\baselineskip} \\ 0.940 / 0.939 / 0.925 / \underline{0.947}}}  & ImageNet-C & 0.988 / 0.986 / 0.981 / \underline{0.989} & 0.929 / 0.928 / 0.894 / \underline{0.932}\\
& ImageNet-R  & 0.979 / 0.984 / 0.972 / \underline{0.984} & 0.923 / 0.926 / 0.886 / \underline{0.927}\\
  & LSUN-C  & \underline{0.994} / 0.990 / 0.989 / 0.993 & 0.938 / 0.931 / 0.904 / \underline{0.940}\\
& LSUN-R  & 0.985 / 0.988 / 0.984 / \underline{0.990} & 0.928 / 0.931 / 0.897 / \underline{0.935}\\
 & iSUN  & 0.985 / 0.989 / 0.984 / \underline{0.991} & 0.928 / 0.932 / 0.896 / \underline{0.936}\\
 \cmidrule{2-4}
 & \textbf{Average}  & 0.986 / 0.987 / 0.982 / \underline{0.989} & 0.929 / 0.930 / 0.895 / \underline{0.934}\\
 \midrule
\midrule
\multirow{6.5}{*}{\shortstack{CIFAR100 \\ \vspace{2\baselineskip} \\ 0.727 / 0.735 / 0.705 / \underline{0.779}}}  & ImageNet-C & 0.886 / 0.929 / 0.925 / \underline{0.930} & 0.641 / 0.686 / 0.652 / \underline{0.696}\\
& ImageNet-R  & 0.815 / 0.910 / \underline{0.934} / 0.920 & 0.572 / 0.674 / 0.658 / \underline{0.687}\\
  & LSUN-C  & \underline{0.967} / 0.931 / 0.901 / 0.965 & 0.685 / 0.680 / 0.643 / \underline{0.751}\\
& LSUN-R  & 0.844 / 0.930 / \underline{0.959} / 0.945 & 0.608 / 0.695 / 0.684 / \underline{0.731}\\
 & iSUN  & 0.842 / 0.923 / 0.954 / \underline{0.955} & 0.603 / 0.689 / 0.680 / \underline{0.734}\\
 \cmidrule{2-4}
 & \textbf{Average}  & 0.871 / 0.925 / 0.935 / \underline{0.943} & 0.621 / 0.685 / 0.663 / \underline{0.720}\\
\bottomrule
\end{tabular}
}
\end{table}

\subsection{Additional Experiments and Discussions}\label{sec4.4}

We further analyzed our BCR method by using various $\lambda$ in our loss function.
Furthermore, we compared our method by formulating another baseline using the triplet loss~\cite{schroff:triplet}.
%and analyzed each component of our class-inclusion loss.
%we explored whether our method maintains the classification accuracy of each vanilla closed-set classifier.
Using the CIFAR10 and TinyImageNet experiments in Setting 1, we present the corresponding OSR results.
We also conducted various additional experiments that can show the effectiveness of our proposed method.

%\vspace{-1\baselineskip}
\paragraph{Selecting $\lambda$.}
Conducting additional OSR experiments with $\lambda \in \{0.1, 0.5, 1, 5, 10\}$ in our loss function $\mathcal{L} = \mathcal{L}_{cf} + \lambda \mathcal{L}_{bg}$,
%where we sampled the original $\mathcal{D}_b$ with the sampling rates of 0.01, 0.1, and 0.5 to simulate various sizes of the background dataset.
Table~\ref{tb:ablation1} presents that our method provides robust OSR accuracy across a \emph{wide range} of $\lambda$, \emph{e.g.}, $\lambda \in [1,5]$, which implies that users can flexibly select $\lambda$ in our method. Although such range may depend on datasets, users are not required to carefully adjust the $\lambda$ parameter.
%$\lambda = 5$ and $\lambda = 1$ showed the best OSR results in the CIFAR10 and TinyImageNet experiments of Setting 1, respectively,
In additional experiments, $\lambda = 5$ yielded the best OSR results in the SVHN and CIFAR + $M$ experiments of Setting 1 and the CIFAR10 experiments of Setting 2,
where $\lambda=0.5$ showed the best results in the CIFAR100 experiments.
%Since the other $\lambda$ values did not significantly degrade OSR performance, one can flexibly select $\lambda$ to balance empirical and open-space risks, where
Such empirical results implies that a lower $\lambda$ value can be better when handling more KKCs.

\begin{table}[!t]
\caption{OSR results with various $\lambda$ in our class-inclusion and the triplet losses. In each cell, the results are presented in the form of
(Accuracy / AUROC / OSCR).}
%\vspace{0.5\intextsep}
\label{tb:ablation1}
\centering
%\fontsize{8.5}{8.5}\selectfont
\resizebox{1\textwidth}{!}{
\begin{tabular}{cC{1.3in}C{1.3in}C{1.3in}C{1.3in}}
\toprule
\multirow{2.5}{*}{Parameter $\lambda$}  & \multicolumn{2}{c}{CIFAR10} & \multicolumn{2}{c}{TinyImageNet}\\
\cmidrule(lr){2-3}\cmidrule(lr){4-5}
& Class inclusion & Triplet & Class inclusion & Triplet \\
\midrule
\midrule
0 (Vanilla)  & 0.963 / 0.759 / 0.472 & -- & 0.785 / 0.631 / 0.308 & -- \\
0.1          & 0.966 / 0.899 / 0.742 & 0.963 / 0.820 / 0.537 & 0.790 / 0.765 / 0.442 & 0.787 / 0.746 / 0.431 \\
0.5          & 0.967 / 0.928 / 0.810 & 0.968 / 0.842 / 0.572 & 0.794 / 0.775 / 0.464 & 0.785 / 0.729 / 0.426 \\
1            & 0.973 / 0.936 / 0.840 & 0.966 / 0.860 / 0.628 & 0.802 / 0.785 / 0.493 & 0.793 / 0.714 / 0.423 \\
5            & 0.973 / 0.948 / 0.870 & 0.965 / 0.872 / 0.628 & 0.798 / 0.783 / 0.480 & 0.785 / 0.707 / 0.360 \\
10           & 0.968 / 0.947 / 0.863 & 0.958 / 0.856 / 0.597 & 0.787 / 0.701 / 0.361 & 0.738 / 0.637 / 0.220 \\
%\midrule
%\textbf{Average} & 0.917 / 0.920 / 0.898 / \textbf{0.931} & 0.900 / 0.899 / 0.885 / \textbf{0.913} & 0.764 / 0.758 / 0.708 / \textbf{0.787} \\
\bottomrule
\end{tabular}
}
\end{table}

%\vspace{-1\baselineskip}
\paragraph{Triplet loss.}
We propose a distance-based BCR method suitable for the OSR problem, where the proposed method defines explicit class-wise boundaries and
then increases the distance gap between KKC and KUC samples based on the boundaries.
Another loss function that can separate KKC and KUC data in terms of such distance measure
is the triplet loss,
where the loss function has been widely employed to control the distances between latent feature vectors effectively.
Therefore, we formulated a baseline distance-based BCR method by following the conventional definition of the triplet loss $\mathcal{L}_{tri}$,
where we set class-wise anchors, KKC training data, and KUC data as
anchors, positive samples, and negative samples, respectively.
Since we observed that training classifiers solely based on the triplet loss $\mathcal{L} = \mathcal{L}_{tri}$ yields significantly worse OSR results
in comparison with the regularization method $\mathcal{L} = \mathcal{L}_{cf} + \lambda \mathcal{L}_{tri}$,
we employed $\mathcal{L}_{tri}$ as a regularization loss function for BCR.
In Table~\ref{tb:ablation1},
we reported experiment results by using the triplet loss as $\mathcal{L}_{bg}$.
The results show that our proposed method (class-inclusion loss) outperforms the regularization method based on the triplet loss.

%\vspace{-1\baselineskip}
\paragraph{Vanilla distance-based classifiers.}
To show the effectiveness of our method, we assessed the OSR performance of vanilla distance-based classifiers (trained solely based on $\mathcal{L}_{cf}$), where we present the results in the form of (Accuracy / AUROC / OSCR).
In the CIFAR10 and TinyImageNet experiments of Setting 1,
we obtained (0.962 / 0.757 / 0.470) and (0.785 / 0.629 / 0.315), respectively.
In the CIFAR10 and CIFAR100 experiments of Setting 2,
the OSR results averaged over the five UUC datasets in vanilla distance-based classifiers were
(0.936 / 0.838 / 0.709) and (0.766 / 0.807 / 0.549), respectively.
Comparing these results to the results in Tables~\ref{tb:table1} and~\ref{tb:table2},
we show that our regularization strategy can significantly improve the OSR performance of distance-based classifiers.

\paragraph{Ablation study on loss terms.}
Recall our loss function $\mathcal{L}_{cf} + \lambda (\mathcal{L}_{bg,k} + \mathcal{L}_{bg,u})$.
As $\mathcal{L}_{bg,u}$ is essential for BCR by making KUC samples located outside explicit class-wise boundaries, we conducted an ablation study to investigate the necessity of $\mathcal{L}_{bg,k}$.
In the absence of $\mathcal{L}_{bg,k}$, which additionally regularizes correctly classified KKC data, we obtained the result of (0.963 / 0.821 / 0.509) for the (Accuracy / AUROC / OSCR) measures in the CIFAR10 experiment of Setting 1, which is worse than our original result (0.973 / 0.948 / 0.870).
This result implies that $\mathcal{L}_{bg,k}$ is necessary to increase the distance gap between KKC and KUC data.

Also, by designing $\mathcal{L}_{bg}$ based on the original HSC loss function (Eq.~(\ref{eq:hscobj})), we obtained the result of (0.950 / 0.634 / 0.338)
for (Accuracy / AUROC / OSCR) in the CIFAR10 experiment of Setting 1,
which supports our hypothesis.

\begin{table}[!t]
\caption{Comparison with the previous OSR methods (Macro-averaged F1 score).}
%\vspace{0.5\intextsep}
\label{tb:table4}
\centering
\resizebox{1\textwidth}{!}{
\begin{tabular}{cC{0.7in}C{0.7in}C{0.7in}C{0.7in}C{0.7in}C{0.7in}C{0.7in}C{0.7in}}
\toprule
\multirow{2.5}{*}{Experiments}  & \multicolumn{4}{c}{Setting 1} & \multicolumn{4}{c}{Setting 2}\\
\cmidrule(lr){2-5}\cmidrule(lr){6-9}
 & SVHN & CIFAR10 & CIFAR+10 & CIFAR+50 & ImageNet-C & ImageNet-R & LSUN-C & LSUN-R \\
\midrule
\midrule
Softmax~\cite{hendrycks:baseline} & 0.725 & 0.600 & 0.701 & 0.637 & 0.639 & 0.653 & 0.642 & 0.647\\
Openmax~\cite{Bendale:openmax} & 0.737 & 0.623 & 0.731 & 0.676 & 0.600 & 0.684 & 0.657 & 0.668\\
CROSR~\cite{Yoshihashi:crosr}   & 0.753 & 0.668 & 0.769 & 0.684 & 0.721 & 0.735 & 0.720 & 0.749\\
CGDL~\cite{sun:cgdl}    & 0.776 & 0.655 & 0.760 & 0.695 & 0.840 & 0.832 & 0.806 & 0.812\\
AOSR~\cite{fang:learning}    & 0.842 & 0.705 & 0.773 & 0.706 & 0.798 & 0.795 & 0.839 & 0.838\\
\midrule
\textbf{Ours}    & \underline{0.854} & \underline{0.761} & \underline{0.805} & \underline{0.732} & \underline{0.876} & \underline{0.869} & \underline{0.880} & \underline{0.877}\\
\bottomrule
\end{tabular}
}
\end{table}

%\vspace{-1\baselineskip}
\paragraph{Previous OSR approaches.}

We additionally compared our proposed approach to previous OSR methods,
whose OSR results are already reported in~\cite{Yoshihashi:crosr,fang:learning}.
For fair comparison, the entire methods presented in Table~\ref{tb:table4} (including ours) were implemented by using a VGG backbone and tested based on the codebase of (\url{https://github.com/Anjin-Liu/Openset_Learning_AOSR}).
The table, which presents OSR results based on
the macro-averaged F1 score measure, shows that our distance-based BCR approach can achieve robust OSR results via a simple inference process in standard classifier architectures.

\paragraph{ResNet-18 architecture.}

In our main experiments, we used the WRN architectures as feature extractors.
To further investigate the effectiveness of our method, we used another standard classifier architecture, ResNet-18~\cite{he2016deep}.
In the first setting, we obtained the quantitative results of {(0.963 / 0.945 / 0.844)}, {(0.966 / 0.950 / 0.845)}, {(0.967 / 0.945 / 0.848)}, and {(0.971 / 0.947 / 0.850)} for the regularization methods using the objectosphere~\cite{Dhamija:reducing}, the uniformity~\cite{Hendrycks:oe}, the energy~\cite{Liu:energy}, and our class-inclusion losses, respectively.
In addition to the results, Table~\ref{tb:table5} shows quantitative results in the second setting, where the results imply that our method can outperform the previous BCR methods with ResNet-18, as we observed in the experiments using the WRN architectures.

%\vspace{-1\baselineskip}
\paragraph{Text classification.}

To show that our proposed BCR method can be applicable in another domain, we compared our class-inclusion loss to the uniformity loss in text classification applications.
For text classification, we used 20 Newsgroups and WikiText103 for KKCs and KUCs, respectively, and trained a simple GRU model~\cite{cho:gru} for $f$ as in~\cite{Hendrycks:oe}.
As UUC sets, we used Multi30K, WMT16, and IMDB.
Since the margin parameters of the objectosphere and the energy losses selected for image classification cannot be suitable for the text classification tasks, we only tested the uniformity loss for comparison.
In Table~\ref{tb:table3}, we present the results, where
we additionally reported the area under the precision-recall curve (AUPR) and the false-positive rate at $95\%$ true-positive rate (FPR95) measures.
As it outperformed the uniformity loss in image classification tasks, our method also showed significantly better OSR accuracy in text classification.
We provide more training details of our text classification models in our supplementary materials.

\begin{table}[!t]
\caption{Comparison with the previous methods in the second setting by using ResNet-18.
The corresponding classification accuracy values are reported
in the first column.}
\label{tb:table5}
\centering
\resizebox{1\textwidth}{!}{
\begin{tabular}{ccC{1.75in}C{1.75in}}
\toprule
 \multirow{2.5}{*}{$\mathcal{D}_{t}/\mathcal{D}_{test}^{k}$} & \multirow{2.5}{*}{$\mathcal{D}_{test}^{u}$} & AUROC ($\uparrow$) & OSCR ($\uparrow$) \\
\cmidrule{3-4}
 & & \multicolumn{2}{c}{Objectosphere / Uniformity / Energy / Class-inclusion (Ours)}\\
\midrule
\midrule
\multirow{6.5}{*}{\shortstack{CIFAR10 \\ \vspace{2\baselineskip} \\ 0.937 / 0.949 / 0.933 / \underline{0.951}}}  & ImageNet-CR     & 0.982 / 0.979 / 0.983 / \underline{0.987} & 0.932 / 0.928 / 0.917 / \underline{0.941}\\
& ImageNet-RE   & 0.977 / 0.982 / 0.975 / \underline{0.988} & 0.918 / 0.934 / 0.909 / \underline{0.945}\\
  & LSUN-CR     & 0.991 / 0.984 / 0.987 / \underline{0.993} & 0.934 / 0.935 / 0.919 / \underline{0.946}\\
& LSUN-RE       & 0.987 / 0.987 / 0.986 / \underline{0.990} & 0.932 / 0.938 / 0.918 / \underline{0.942}\\
 & iSUN         & 0.987 / 0.986 / 0.987 / \underline{0.994} & 0.932 / 0.938 / 0.919 / \underline{0.946}\\
 \cmidrule{2-4}
 & \textbf{Average}  & 0.985 / 0.984 / 0.984 / \underline{0.991} & 0.930 / 0.935 / 0.916 / \underline{0.944}\\
\bottomrule
\end{tabular}
}
\end{table}

\begin{table}[!t]
\caption{Comparison with the previous BCR method in text classification experiments.}
\label{tb:table3}
%\vspace{0.5\intextsep}
\centering
%\fontsize{8.5}{8.5}\selectfont
\resizebox{0.83\textwidth}{!}{
\begin{tabular}{ccC{0.85in}C{0.85in}C{0.85in}C{0.85in}}
\toprule
 \multirow{2.5}{*}{$\mathcal{D}_{t}/\mathcal{D}_{test}^{k}$} & \multirow{2.5}{*}{$\mathcal{D}_{test}^{u}$} & AUROC ($\uparrow$) & AUPR ($\uparrow$) & FPR95 ($\downarrow$) & OSCR ($\uparrow$)\\
\cmidrule{3-6}
 & & \multicolumn{4}{c}{Uniformity / Class-inclusion (Ours)}\\
\midrule
\midrule
\multirow{4.5}{*}{\shortstack{20 Newsgroups \\ \vspace{2\baselineskip} \\ 0.719 / \underline{0.749}}}  & Multi30k &  0.997 / \underline{0.997} &  \underline{0.998} / 0.997 &  \underline{0.002} / 0.010&  0.715 / \underline{0.745}\\
& WMT16 & \underline{0.997} / 0.996 & \underline{0.997} / 0.995 & \underline{0.010} / 0.016&  0.715 / \underline{0.742}\\
& IMDB &  0.805 / \underline{0.999} &  0.692 / \underline{0.999} &  0.367 / \underline{0.003}&  0.585 / \underline{0.747}\\
 \cmidrule{2-6}
 & \textbf{Average} &  0.933 / \underline{0.997} &  0.896 / \underline{0.997} &  0.126 / \underline{0.010}&  0.672 / \underline{0.745}\\
\bottomrule
\end{tabular}
}
\end{table}

\section{Concluding Remarks}

In this paper, we propose a novel BCR method to train open-set classifiers that can provide robust OSR results with a simple inference process.
By employing distance-based classifiers with the principle of LDA, we designed a novel class-inclusion loss based on the principle of probability of inclusion,
which effectively limits the feature space of KKC data in a class-wise manner and then regularizes KUC samples to be located far away from the limited class-wise spaces.
Through our extensive experiments, we present that our method can achieve robust UUC rejection performance, while maintaining high closed-set classification accuracy.
As this paper aims to improve the reliability of modern DNN-based classifiers,
we hope our work to enhance reliability and robustness in various classification applications by providing a novel methodology of handling UUC samples.
%\vspace{0.4\baselineskip}\\
%{\bf Future works} \quad
%When designing our class-inclusion loss for distance-based classifiers, we exploit the class-conditioned Gaussian assumption of LDA by assuming identity covariance matrices.
%For future works, one can integrate our proposed method with generative models that can build a class-conditioned Gaussian feature space or employ additional regularization or training methods that can learn a better formulation for covariance matrices.
%Furthermore, KUC data generation or resampling techniques can be another line of future research to acquire a compact and effective $\mathcal{D}_b$.

%\subsubsection{Limitations.}
%This work was supported by Institute of Information \& communications Technology Planning \& Evaluation (IITP) grant funded by the Korea government (MSIT)  (No.2019-0-00075, Artificial Intelligence Graduate School Program (KAIST)).

\subsubsection{Acknowledgements.}
This work was supported by Institute of Information \& communications Technology Planning \& Evaluation (IITP) grant funded by the Korea government (MSIT) (No.2019-0-00075, Artificial Intelligence Graduate School Program (KAIST)) and the National Research Foundation of Korea (NRF) grants funded by the Korea government (MSIT) (No. NRF-2018M3E3A1057305 and No. NRF-2022R1A2B5B02001913).

\clearpage
\bibliographystyle{splncs04}
\bibliography{egbib}
\end{document}